\pdfoutput=1

\documentclass[11pt]{article}

\usepackage[preprint]{acl}

\usepackage{times}
\usepackage{latexsym}

\usepackage[T1]{fontenc}

\usepackage[utf8]{inputenc}

\usepackage{microtype}

\usepackage{inconsolata}

\usepackage{graphicx}

\usepackage{times}
\usepackage{latexsym}
\usepackage[T1]{fontenc}     
\usepackage[utf8]{inputenc}  
\usepackage{microtype}       
\usepackage{inconsolata}     
\usepackage{url}             
\usepackage{amsmath}         
\usepackage{graphicx}        
\usepackage{booktabs}
\usepackage{multirow}
\usepackage{tabularx}
\usepackage{stfloats}
\usepackage{caption}
\usepackage{natbib}

\title{Evaluating the Effectiveness of Black-Box Prompt Optimization as the Scale of LLMs Continues to Grow}

\author{Ziyu Zhou\textsuperscript{1}, Yihang Wu\textsuperscript{1}, Jingyuan Yang\textsuperscript{1}, Zhan Xiao\textsuperscript{2}, Rongjun Li\textsuperscript{1}, \\
\textsuperscript{1}IT Innovation and Research Center, Huawei Technologies, \textsuperscript{2}Zhejiang Lab \\
\{zhouziyu8,, yangjingyuan2, lirongjun3\}@huawei.com\\ wuyihang2@h-partners.com, xz@zhejianglab.org}

\begin{document}
\maketitle

\begin{abstract}
Black-Box prompt optimization methods have emerged as a promising strategy for refining input prompts to better align large language models (LLMs), thereby enhancing their task performance. Although these methods have demonstrated encouraging results, most studies and experiments have primarily focused on smaller-scale models (e.g., 7B, 14B) or earlier versions (e.g., GPT-3.5) of LLMs. As the scale of LLMs continues to increase, such as with DeepSeek V3 (671B), it remains an open question whether these black-box optimization techniques will continue to yield significant performance improvements for models of such scale. In response to this, we select three well-known black-box optimization methods and evaluate them on large-scale LLMs (DeepSeek V3 and Gemini 2.0 Flash) across four NLU and NLG datasets. The results show that these black-box prompt optimization methods offer only limited improvements on these large-scale LLMs. Furthermore, we hypothesize that the scale of the model is the primary factor contributing to the limited benefits observed. To explore this hypothesis, we conducted experiments on LLMs of varying sizes (Qwen 2.5 series, ranging from 7B to 72B) and observed an inverse scaling law, wherein the effectiveness of black-box optimization methods diminished as the model size increased.
\end{abstract}

\section{Introduction}
Prompt optimization methods have emerged as an effective strategy for enhancing task performance by carefully refining input prompts to better align with LLMs \citep{Brown2020LanguageModels}. Broadly speaking, existing prompt optimization methods can be classified into two categories: white-box and black-box prompt optimization methods. White-box prompt optimization techniques typically involve utilizing gradient information to refine prompts. For instance, AutoPrompt \citep{shin2020autoprompt} uses gradient-based methods to iteratively replace discrete prompt tokens, refining the initial prompt and improving performance on downstream tasks. Similarly, prefix tuning \citep{liu2022p} and prompt tuning \citep{lester2021power} fine-tune additional soft continuous embeddings, referred to as "soft tokens," to construct more effective task-specific prompts. Although these methods show promising results, they require access to the model's internal gradients or parameters, limiting their applicability in many closed-source models, such as GPT4o \citep{Hurst2024GPT4o} and Gemini \citep{Anil2023GeminiModels}.

Another category of prompt optimization methods is based on nonparametric black-box techniques. These methods typically optimize prompts through calling external APIs, without the need to access the internal model parameters or gradients. For example, EvoPrompt \citep{Guo2023Connecting} utilizes evolutionary algorithms to iteratively search for better task prompts through crossover and mutation. Similarly, methods like ProTeGi \citep{pryzant2023automatic}, BPO \citep{Cheng2023Black}, and OPRO \citep{Yang2023Large} use LLMs themselves as optimizers, generating improved task prompts by leveraging text feedback signals from the LLMs. Despite these methods demonstrating substantial performance improvements, they have primarily been tested on smaller-scale LLMs (e.g., those with fewer than 14B parameters) or earlier versions of LLMs (e.g., GPT-3.5 \citep{Ye2023GPT}). As LLMs continue to scale up, it is still uncertain whether these black-box optimization techniques will maintain their ability to deliver substantial performance gains.

To address this question, we selected three popular black-box optimization methods and evaluated their performance on large-scale LLMs, DeepSeek V3 \citep{DeepSeek2024V3} and Gemini 2.0 Flash \citep{GoogleGemini2024}, across four NLU and NLG benchmark datasets. The experimental results demonstrate that the performance improvements from these methods have become less significant. For the NLU datasets, the average accuracy improvements for DeepSeek V3 and Gemini 2.0 Flash across these three  optimization methods were 0.86\% and 1.16\%, respectively. Similarly, for the NLG datasets, the corresponding metric improvements for DeepSeek V3 and Gemini 2.0 Flash were 1.04\% and 2.03\%, respectively. We hypothesize that the limited improvements are primarily due to the issue of model scale. To investigate this further, we conducted experiments on LLMs of varying sizes, specifically the Qwen 2.5 series, with model sizes ranging from 7B to 72B parameters. The results revealed an inverse scaling law, in which the efficacy of black-box optimization methods decreased as the model size increased. In brief, our work offers two key contributions:

\begin{itemize}
    \item We evaluate three black-box optimization methods on large-scale LLMs  using four NLU and NLG datasets, finding only limited improvements in performance.
    \item Our findings reveal an inverse scaling pattern, where the effectiveness of black-box optimization decreases as the size of the LLM increases.
\end{itemize}

\section{Related Work}
\subsection{White-Box Prompt Optimization Methods}
Early white-box prompt optimization methods, such as AutoPrompt \citep{shin2020autoprompt}, utilize gradients to search for discrete prompt tokens to improve model performance. \citet{wen2023hard} expanded these hard prompt optimization methods to multimodal tasks, including text-to-image generation. Prefix-Tuning \citep{li2021prefix} introduced continuous, task-specific vectors as ``soft tokens," optimizing them via gradients to boost performance. Furthermore, P-Tuning v2 \citep{liu2022p} optimized ``soft embeddings" across multiple transformer layers, achieving improvements across a broader range of tasks. More recently, GReaTer \citep{das2024greater} incorporated reasoning path information into gradient-based prompt searches, yielding significant performance improvements over prior methods.

\subsection{Black-Box Prompt Optimization}
Black-box prompt optimization methods seek to enhance task performance by refining prompts without accessing the model's internal parameters or gradients. For example, EvoPrompt \citep{Guo2023Connecting} employs evolutionary algorithms, including crossover and mutation, to iteratively refine prompts. APE \citep{Zhou2022Large} frames black-box prompt optimization as a program synthesis problem, refining prompts through top-k sampling and resampling. OPRO \citep{Yang2023Large} integrates historical optimization trajectory information to improve the stability of the optimization process. ProTeGi \citep{pryzant2023automatic} refines prompts through iterative language feedback, resulting in enhanced performance. Likewise, BPO \citep{Cheng2023Black} optimizes prompts using human feedback and utilizes a small LLM as a prompt optimizer, reducing the high costs associated with large-scale LLMs.

\begin{table*}[!htbp]\footnotesize
\centering
\begin{tabularx}{0.85\textwidth}
{@{\extracolsep{\fill}}lcccc}
\toprule
\textbf{Model} & \textbf{SST-5 (acc.)} & \textbf{AG's News (acc.)} & \textbf{SAMSum (ROUGE)} & \textbf{ASSET (SARI)} \\
\midrule
\multicolumn{5}{l}{\textbf{DeepSeek V3}} \\
\midrule
+ EvoPrompt & 56.0 $\rightarrow$ 56.6 & 83.6 $\rightarrow$ 84.8 & 34.4 $\rightarrow$ 35.4 & 45.3 $\rightarrow$ 45.8 \\
+ ProTeGi & 56.0 $\rightarrow$ 56.4 & 84.0 $\rightarrow$ 85.8 & 33.9 $\rightarrow$ 33.7 & 46.4 $\rightarrow$ 46.9 \\
+ BPO & 56.0 $\rightarrow$ 56.4 & 84.6 $\rightarrow$ 83.8 & 33.9 $\rightarrow$ 34.1 & 45.3 $\rightarrow$ 45.8 \\
Average \% Increase & \textbf{0.83\%} & \textbf{0.88\%} & \textbf{0.97\%} & \textbf{1.10\%} \\
\midrule
\multicolumn{5}{l}{\textbf{Gemini 2.0 Flash}} \\
\midrule
+ EvoPrompt & 56.4 $\rightarrow$ 56.8 & 82.4 $\rightarrow$ 85.4 & 37.2 $\rightarrow$ 38.5 & 45.4 $\rightarrow$ 47.6 \\
+ ProTeGi & 55.6 $\rightarrow$ 56.2 & 82.5 $\rightarrow$ 83.5 & 37.2 $\rightarrow$ 37.6 & 44.6 $\rightarrow$ 46.0 \\
+ BPO & 57.6 $\rightarrow$ 58.2 & 82.8 $\rightarrow$ 82.2 & 37.2 $\rightarrow$ 36.9 & 44.2 $\rightarrow$ 44.4 \\
Average \% Increase & \textbf{0.94\%} & \textbf{1.38\%} & \textbf{1.25\%} & \textbf{2.81\%} \\
\bottomrule
\end{tabularx}
\caption{Performance of Black‑Box Prompt Optimization Methods on DeepSeek V3 \& Gemini 2.0 Flash.}
\label{tab:main}
\end{table*}

\begin{table*}[ht]
    \centering
    \renewcommand{\arraystretch}{1.2}
    \small  

    \begin{tabularx}{\textwidth}{l X}
        \toprule
        \textbf{Model} & \textbf{Comparison of the Initial and Optimized Prompts on the AG's News Dataset} \\
        \midrule
        Initial          & Identify the category of the text (e.g.\ Technology, Sports, World, Business). \\
        DeepSeek V3       & Identify the \textbf{main topic} of the content and \textbf{select} from the categories: World, Sports, Business, or Tech. \\
        Gemini 2.0 Flash  & \textbf{Categorize} the following \textbf{news article} under one of these themes: World, Sports, Business, or Tech. Identify the article's \textbf{primary subject} to make your selection. \\
        \midrule
        \textbf{Model} & \textbf{Comparison of the Initial and Optimized Prompts on the SAMSum Dataset} \\
        \midrule
        Initial          & Please summarize the main context. \\
        DeepSeek V3       & \textbf{Provide} a \textbf{clear and concise} summary of the main idea, \textbf{removing} any \textbf{redundant} or \textbf{extraneous information}. \\
        Gemini 2.0 Flash  & \textbf{Create} a \textbf{very short, jargon‑free} summary that \textbf{captures} the \textbf{core message} and \textbf{vital information}, \textbf{avoiding} any \textbf{repetition} or \textbf{fluff}. \\
        \bottomrule
    \end{tabularx}

    \caption{Comparison of the Initial and Optimized Prompts on DeepSeek V3 and Gemini 2.0 Flash.}
    \label{tab:llm_comparison}
\end{table*}

\section{The Effectiveness of Black-Box Prompt Optimization Methods on Large-Scale LLMs}
\label{main exp}
\subsection{Datasets and Evaluation Metrics}
The four datasets used in this study include SST-5 \citep{Socher2013Recursive}, a dataset for sentiment classification based on movie reviews; AG's News \citep{Zhang2015CharacterLevel}, a corpus for news categorization across four primary topics: World, Sports, Business, and Sci/Tech; SAMSum \citep{Gliwa2019SAMSum}, a dialogue summarization using messenger-style conversations and ASSET \citep{Alva-Manchego2020ASSET}, a dataset for sentence simplification, where each sentence is paired with multiple reference simplifications. For NLU datasets, we randomly sample 500 examples as training dataset for prompt optimization and 500 examples as test dataset for evaluation, while the NLG datasets are trained and assessed on their complete examples respectively. For evaluation metrics, accuracy is used for SST-5 and AG's News, while ROUGE-L \citep{Lin2004ROUGE} and SARI \citep{xu-etal-2016-optimizing} are employed for SAMSum and ASSET, respectively.

\subsection{Experimental Design}
Three black-box prompt optimization methods are utilized for evaluation. Specifically, the EvoPrompt method \citep{Guo2023Connecting} refines the initial prompts through a stepwise evolutionary process, generating candidate prompts via crossover and mutation, and selecting the best‑performing prompt after four iterative optimization cycles on the training data. The ProTeGi method \citep{pryzant2023automatic} optimizes initial prompts by leveraging text language gradients derived from the training data, also undergoing four optimization rounds. The BPO method \citep{Cheng2023Black} directly applies the released sequence-to-sequence prompt optimizer, performing five optimization rounds. For all three black-box prompt optimization methods, we evaluate them on these four datasets using large-scale LLMs, including DeepSeek V3 \citep{DeepSeek2024V3} and Gemini 2.0 Flash \citep{GoogleGemini2024}.

\subsection{Results and Analysis}
As presented in Table \ref{tab:main}, Black-Box prompt optimization methods show limited improvements in performance when applied to larger scale LLMs. Specifically, for DeepSeek V3, the average improvement across NLU tasks was only 0.86\%, and 1.16\% for NLG tasks. Similarly, for Gemini 2.0 Flash, the NLU task improvement was 1.04\%, and the NLG task improvement was 2.03\%. These results suggest that prompt optimization has a minimal effect on very large models. To explore this further, we conducted a comparative analysis of prompts before and after optimization using the EvoPrompt method. As shown in Table \ref{tab:llm_comparison}, the optimized prompts exhibit only slight modifications compared to the initial prompts for both datasets. The primary adjustments involve replacing synonyms and subtly rephrasing to improve clarity. For instance, in the SAMSum dataset, the initial prompt simply instructs, “Please summarize the main context.” After optimization, the prompts become more detailed, such as “Provide a clear and concise summary of the main idea, removing any redundant or extraneous information.” (DeepSeek V3), or “Create a very short, jargon-free summary that captures the core message and vital information, avoiding any repetition or fluff.” (Gemini 2.0 Flash). These minor synonym substitutions are unlikely to have a significant impact on large-scale LLMs. This could be because, generally, larger LLMs exhibit more refined alignment, making them less sensitive to such subtle variations in lexical choices. Similar findings are discussed in \citep{Shirafuji2023Exploring}, where the authors explore the effects of superficial prompt changes in code generation tasks.

\section{The Impact of LLMs Scale for Black-Box Prompt Optimization}
\subsection{Experimental Design}
To examine whether the size of an LLM influences the effectiveness of black‑box prompt optimization, we evaluated the Qwen‑2.5 family, encompassing models from 7B to 72B parameters. Specifically, we applied the EvoPrompt black‑box optimization method under the same experimental setup described in Section ~\ref{main exp}.

\begin{figure*}[ht]
    \centering
    \includegraphics[width=\linewidth, keepaspectratio]{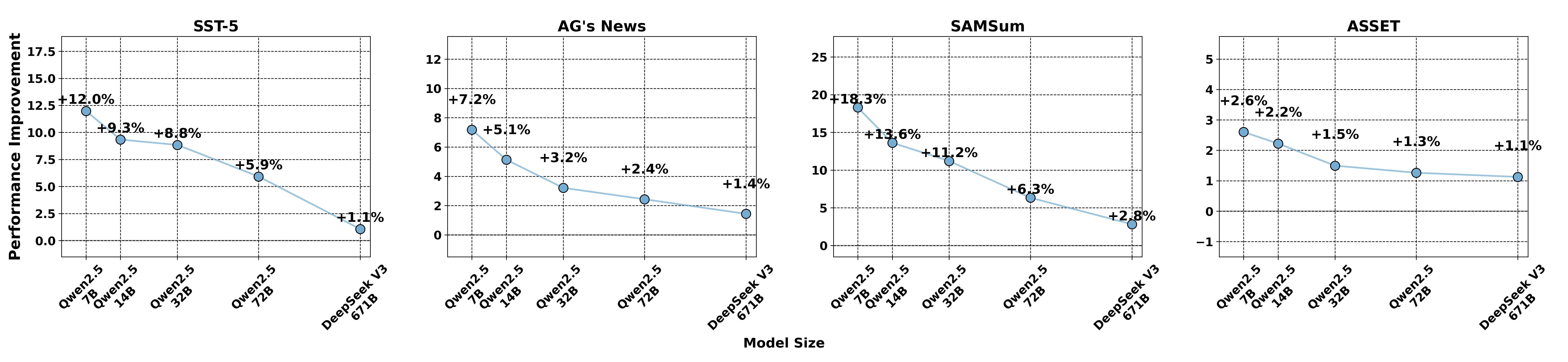}
    \caption{Performance Improvements of EvoPrompt Across Different Scales of Qwen 2.5 Series.}
    \label{fig:abl_111111}
\end{figure*}

\begin{table*}[ht]
    \centering
    \renewcommand{\arraystretch}{1.2}
    \small  
    \begin{tabularx}{\textwidth}{l X}
        \toprule
        \textbf{Model} & \textbf{Comparison of the Initial and Optimized Prompts on AG's News Dataset.} \\
        \midrule
        Initial          & Identify the category of the text (e.g.\ Technology, Sports, World, Business). \\
        Qwen2.5 7B        & Identify the \textbf{main topic} of the \textbf{news article} and \textbf{classify} it under "World", "Sports", "Tech", or "Business". \\
        Qwen2.5 14B       & Based on the \textbf{primary theme} of the provided \textbf{news article}, \textbf{categorize} it as "World", "Sports", "Tech", or "Business". \\
        Qwen2.5 32B       & \textbf{Determine} the \textbf{primary topic} of the \textbf{news article} and \textbf{select} from the following categories: World, Sports, Business, or Tech. \\
        Qwen2.5 72B       & Your task is to identify the \textbf{primary topic} of the \textbf{news article} and \textbf{choose} from World, Sports, Business and Tech. \\
        \midrule
        \textbf{Model} & \textbf{Comparison of the Initial and Optimized Prompts on SAMSum Dataset.} \\
        \midrule
        Initial          & Please summarize the main context. \\
        Qwen2.5 7B        & \textbf{Use concise language} to summarize the main points, \textbf{avoiding} any \textbf{unnecessary details} or \textbf{repetition}. \\
        Qwen2.5 14B       & Summarize the main context \textbf{briefly}, \textbf{focusing} only on the \textbf{key points} and \textbf{omitting} any \textbf{redundant} or \textbf{irrelevant information}. \\
        Qwen2.5 32B       & Summarize the \textbf{key points briefly}, \textbf{omitting} any \textbf{extraneous details} or \textbf{repetition}. \\
        Qwen2.5 72B       & \textbf{Use concise language} to summarize the key points, \textbf{ensuring clarity} and \textbf{omitting unnecessary details} or \textbf{repetition}. \\
        \bottomrule
    \end{tabularx}

    \caption{Comparison of the Initial and Optimized Prompts across different Qwen 2.5 Scales.}
    \label{tab:prompt_comparison}
\end{table*}

\subsection{Results and Analysis}
Figure \ref{fig:abl_111111} illustrates a distinct inverse-scaling phenomenon, wherein the improved performance gains obtained from black-box prompt optimization methods decline significantly as model scale increases. Specifically, on the SST-5 benchmark, accuracy improvements diminish notably from 12\% for the Qwen-2.5 7B model to 5.9\% for the Qwen-2.5 72B model, ultimately reaching just 1.1\% for the DeepSeek-V3 671B model. Comparable trends are observed across other datasets.

To further investigate the underlying reasons for these observations, we analyzed performance gains across the Qwen 2.5 series. As illustrated in Table \ref{tab:prompt_comparison}, smaller LLMs (7B and 14B) exhibit a relative significant improvement, likely attributable to the incorporation of domain-specific clues in the optimized prompts. For instance, in the case of AG's News, the optimized prompt explicitly includes the phrase ``news article'', providing clear, context-specific guidance that smaller models greatly benefit from. 

Meanwhile, larger models (32B and 72B) yield relatively modest improvements. This may be due to the fact that larger models inherently possess a more comprehensive domain understanding and semantic alignment, making explicit domain cues gradually redundant, while lexical refinements or synonym replacements, such as "identify" to "determine," become ineffective.

\section{Conclusion}
In this paper, we investigate whether black‑box prompt optimization can deliver substantial benefits for large‑scale LLMs. Our experiments reveal that as model size increases, the performance gains on both NLU and NLG datasets progressively diminish, exhibiting a clear inverse scaling trend. 

\section*{Limitations}
Our preliminary experiments indicate that black‑box prompt optimization yields only limited benefits for large‑scale LLMs. Nonetheless, several limitations temper the scope of our conclusions. First, the largest Qwen‑2.5 model we assess contains 72B parameters, leaving an unexplored gap between this scale and the 671B DeepSeek‑V3, intermediate‑sized models therefore remain untested. Second, our analyses focus on English‑language benchmarks, restricting the generalizability of the findings to multilingual contexts, especially low‑resource languages, whose response to prompt optimization is still unknown. Third, we only consider text‑based prompts, leaving multi‑modal prompt optimization, incorporating visual or audio modality unexamined. Furthermore, our evaluation omits treasoning‑oriented LLMs, such as DeepSeek R1 \citep{DeepSeek2025R1} or OpenAI o3 \citep{OpenAI2025o3_o4_mini_systemcard}, which may display distinct scaling behavior and prompt‑sensitivity characteristics.

\bibliography{custom}  


\end{document}